\documentclass[lettersize, journal, 10pt]{IEEEtran}
\IEEEoverridecommandlockouts
\usepackage{cite}
\usepackage{amsmath,amssymb,amsfonts}
\usepackage{algorithmic}
\usepackage{algorithm}
\usepackage{array}
\usepackage[caption=false,font=normalsize,labelfont=sf,textfont=sf]{subfig}
\usepackage{graphicx}
\usepackage{textcomp}
\usepackage{xcolor}
\usepackage{url}
\usepackage{verbatim}
\usepackage{cite}
\usepackage{stfloats}

\usepackage{multirow}
\usepackage{multicol}
\usepackage{comment}

\begin{document}

\title{An On-Chip Trainable Neuron Circuit for SFQ-Based Spiking Neural Networks}

\author{\IEEEauthorblockN{Beyza Zeynep Ucpinar, Mustafa Altay Karamuftuoglu, Sasan Razmkhah, Massoud Pedram}\\
\IEEEauthorblockA{\textit{Ming Hsieh Department of Electrical and Computer Engineering} \\
\textit{University of Southern California, Los Angeles, USA}}
}
\maketitle

\noindent
\begin{abstract}
We present an on-chip trainable neuron circuit. Our proposed circuit suits bio-inspired spike-based time-dependent data computation for training spiking neural networks (SNN). The thresholds of neurons can be increased or decreased depending on the desired application-specific spike generation rate. This mechanism provides us with a flexible design and scalable circuit structure. We demonstrate the trainable neuron structure under different operating scenarios. The circuits are designed and optimized for the MIT LL SFQ5ee fabrication process. Margin values for all parameters are above 25\% with a 3GHz throughput for a 16-input neuron.
\end{abstract}

\section{Introduction}
\noindent
Neuromorphic computing is the foundation of deep learning and artificial intelligence (AI) that draws inspiration from the structure and functioning of the human brain \cite{Furber_2016}. Deep neural networks (DNNs) have proven to be an excellent model for learning systems. However, the training of such networks can be time and energy-consuming. One class of DNNs, known as the Spiking neural networks (SNN), are highly compatible with the biological brain regarding their learning style and use of discrete spikes for information transfer between neurons \cite{snn_brain}, \cite{Roy2019TowardsSM}. Single Flux Quantum (SFQ) logic \cite{likharevRSFQ, razmkhahBook} also uses discrete spike-like pulses for computing. Therefore, with orders of magnitude lower power and higher speed than state-of-the-art CMOS, SFQ is a good candidate for implementing SNN architecture. 

The neuron is the core part of neural networks.\cite{neuron_def} A neuron circuit includes an accumulator and a threshold unit. There are several works implementing neuron circuit design with superconductors, primarily focused on inference applications\cite{}. However, on-chip trainability, especially for SNN, is an important feature that needs to be developed.

Traditional Artificial Neural Networks (ANN) primarily rely on continuous-valued activations to perform neuromorphic computing. However, the human brain, comprising billions of interconnected neurons communicating through synapses, employs discrete spikes or pulses for communication \cite{Yang2020NeuromorphicEF}. SNN adopts a more brain-like approach, where neurons communicate through discrete pulses. These spiking pulses encode the timing and frequency of neuronal activations, allowing for more biologically plausible computations \cite{Hendy2022ReviewOS}. One remarkable aspect of neuromorphic computing is its event-driven processing methodology. Unlike conventional computing systems that process data continuously, neuromorphic systems respond exclusively to significant changes or events in the input data. This event-driven approach significantly reduces overall computation time and, more importantly, leads to exceptional energy efficiency.
Even though SNN shows great promise \cite{intel1}, \cite{ibm2}, SNN training, especially on-chip training, is challenging. On-chip training of SNNs usually yields lower accuracy than ANNs due to its forward learning approach. 

The neuron in an SNN plays a crucial role in information processing and learning. It has a thresholding unit in it, which is called soma. When the summation of the inputs exceeds the threshold value, the neuron fires and generates an SFQ pulse; otherwise, the neuron remains silent as described in equation \ref{eq:neuron1}. 
\begin{equation}
\mathcal{O}_i = 
\begin{cases} 
1, & \sum_{k=1}^{N} w_{k} \times x_{k} \geq T_{i}\\
0, & \sum_{k=1}^{N} w_{k} \times x_{k} < T_{i}
\end{cases}
\label{eq:neuron1}
\end{equation}
where the $w_{k}$ is the weight parameter, $x_{k}$ denotes the input value of the $i^{th}$ neuron, and $T_{i}$ is the neuron's threshold value. 
The threshold value is a significant parameter, assuming different values depending on the application. Lower threshold values render neurons more sensitive to inputs, while higher threshold values reduce their sensitivity. The threshold value is kept constant in conventional SNNs. However, the ability to adapt the threshold value can substantially enhance the network performance \cite{strain}.

The abaility to change a neuron's threshold value in an on-chip network offers several benefits \cite{hornd,xudong}. Firstly, it confers flexibility and adaptability to the network. Different tasks or stages in a neural network may necessitate varying sensitivity to incoming signals, implying the requirement for different thresholds. By adjusting the threshold value of individual neurons, neurons within the same layers, or neurons within the same kernels, the network's behavior can be finely tuned to be more or less sensitive to specific input patterns. Diehlet et al. \cite{diehl} claim that an adaptive membrane threshold mechanism must be employed to prevent single neurons from dominating the response pattern. With the proposed architecture, the authors increased the accuracy from 93\% to 95\%. This becomes more important during training to avoid a new data class dominating the network and overwriting the previous training. Reference \cite{zhong} shows the occurrence of overfitting can be effectively suppressed by using an adaptive threshold. It goes on to show that the number of excitation pulses decreases with the help of an adaptive threshold. Hence, it is advantageous when optimizing the energy consumption of the chip. The authors report 96\% accuracy on the MNIST dataset with the provided method. Reference \cite{shaban} proposes an adaptive threshold neuron method that offers fast convergence, higher accuracy, and flexibility. They reached 96.1\% accuracy on the SMIST dataset. Reference \cite{chen} claims that the ratio of threshold to weights (RTTW), the balance between weight and threshold values, affects the accuracy. With the adaptive threshold method, they achieved 93.93\% accuracy. Such adaptability is also advantageous when optimizing the energy consumption of the neural network on the chip. Appropriate threshold values enable the reduction of unnecessary computations, leading to lower dynamic power consumption. 

This work introduces a novel feature for SNNs: adjustable neuron thresholds. These thresholds can be modified individually during training or for specific inference networks, ensuring high-margin values. The adjustable threshold structure has a footprint of $\mathrm{120\times90\mu m^2}$ for the Threshold Adjustment Unit (TAU)  and $\mathrm{60\times30\mu m^2}$ for each Threshold Unit (TU) with a fixed threshold value of 2. The threshold adjustment time is only 40 ps due to the circuit's synchronous nature.
\vspace{-2mm}

\section{Methodology}
\noindent
The neuron circuit is a combination of both TAU, TU, and Arbiter. A predetermined, hardware-assigned threshold value, a significant parameter in the system, characterizes the TAU. Also, the threshold ceiling of the TAU defines the uppermost achievable threshold value, and this limit is an even number. The TA unit loads the initial data depending on the desired threshold value and changes the ground state of the TU. This operation is akin to introducing a bias level into a system. The TA has the increment and decrement information and generates the load data for changing the TU's internal states and the system's threshold value. The Arbiter unit merges the load data from the TA Unit and the input data. To simplify its functionality for conceptual clarity, it can be likened to a lossless Confluence Buffer Unit (CBU). Essentially, the Arbiter unit combines the load and input data and subsequently conveys this combined information to the set input of the TU. However, the CBU may introduce an undesirable loss of SFQ pulses depending on the timing requirements imposed on its input signals. We used the Arbiter to solve this issue. The distinctive characteristic of the Arbiter is its capacity to merge data without causing any loss of SFQ pulses, addressing a critical concern in the system's performance and reliability. 

\begin{figure}[!h]
\centering
  \includegraphics[width=0.7\linewidth]{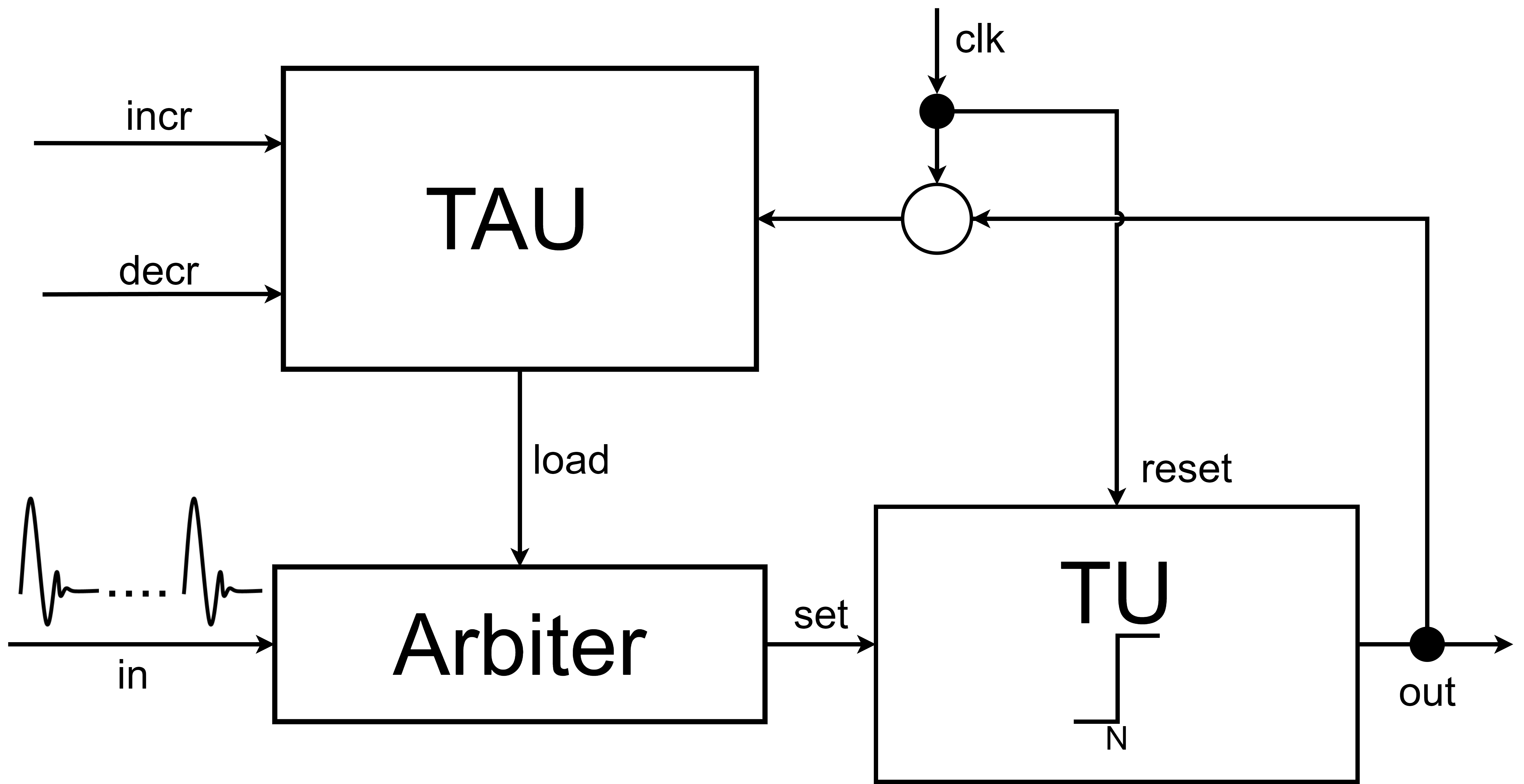}
  \caption{Neuron Circuit Block Diagram with TAU, TU, and arbiter. The TA has increment and decrement pins that adjust the load value. The arbiter then applies this load value with the input signals to the TU, generating the output. Each output triggers the TAU to reload the data to the arbiter. 
}
  \label{fig:systembd}
  \vspace{-2mm}
\end{figure}

The block diagram of the Neuron Circuit is given in Fig. \ref{fig:systembd}.

The adjusted threshold value is calculated as follows:
Adjusted\_Threshold $=$ Max\_Threshold $-$ Load.
The hardware determines the Maximum Threshold Value, and the Load Value comes from the TAU. If the purpose is to change the threshold values layer by layer or kernel by kernel, then the increment and decrement pins of the neurons in the same layer can be connected. 
\vspace{-2mm}

\section{Circuit Implementations}
\vspace{-1mm}
\subsection{Thresholding Unit}
\begin{figure}[!htb]
\centering
  \includegraphics[width=0.8\linewidth]{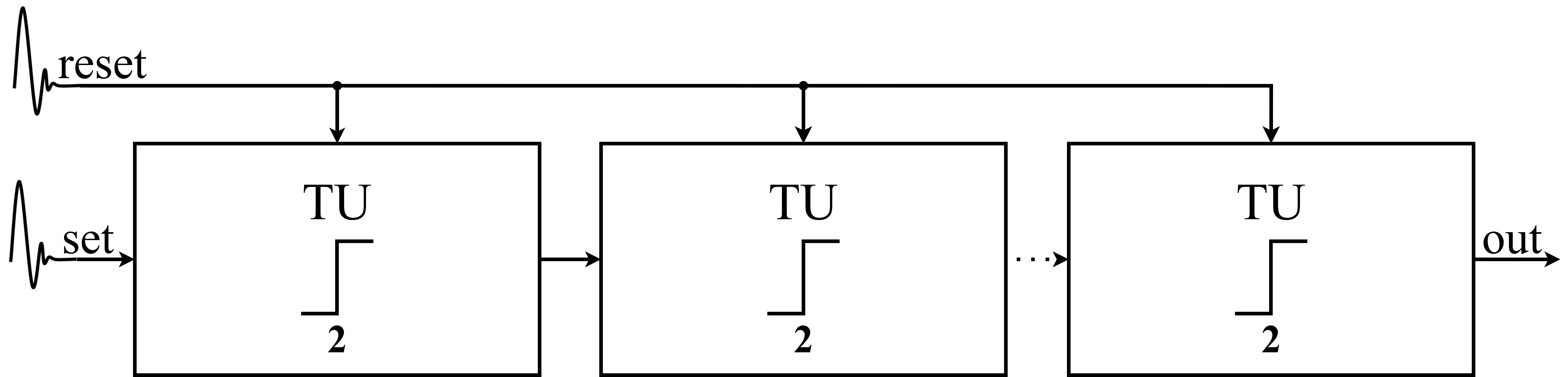}
  \caption{Threshold Unit Cascading Structure. The threshold unit consists of a series of RTFFs. Adding one RTFF increases the maximum threshold by two.}
  \label{fig:tu_multi}
  \vspace{-2mm}
\end{figure}

\noindent
The proposed circuit includes three key components. The first component, TU, operates asynchronously and exhibits high-speed characteristics. The structure employed for thresholding is based on Toggle Flip Flop (TFF). TFF can be likened to a frequency divider. The circuit itself has two states: S1 and S2. The idle state is S1, and the first input changes the state from S1 to S2. When the second input comes, it generates an SFQ pulse and returns to the S1 state. However, it resets the state machine whenever the reset arrives, and the circuit starts from S1. It has a scalable nature, wherein the cascading of two TFFs results in the division of frequency twice. Consequently, a single TFF suffices for implementing a threshold of 2, while two TFFs are utilized for achieving a threshold of 4, three TFFs are used for performing a threshold of 6, and so on. The minimum achievable threshold value with the TFF structure is 2. Thanks to its asynchronous nature, the throughput is limited by the TFFs recovery time, which is in order of 100GHz. The block diagram of the TU and its cascading structure is given in Fig. \ref{fig:tu_multi}. The reset signal is the same for all TUs.

\begin{figure}[htp]
    \centering
    \begin{minipage}[t]{0.45\columnwidth}
        \centering
        \includegraphics[width=\linewidth]{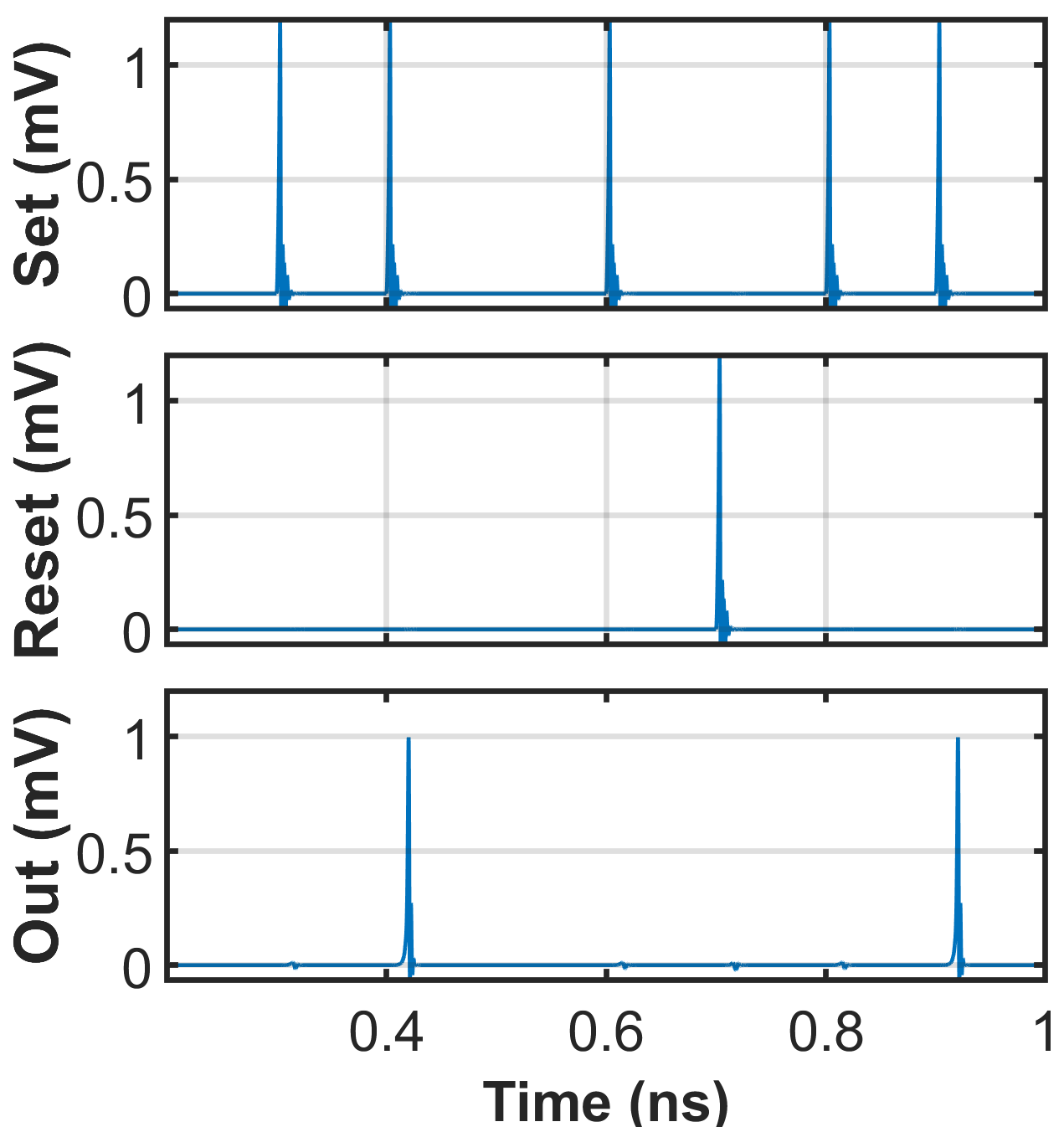}
        \captionsetup{font=small}
        \caption{Simulations result of a TU with one RTFF, which means the threshold value of two.}
        \label{fig:sim_th2}
        \vspace{-2mm}
    \end{minipage}%
    \hfill
    \begin{minipage}[t]{0.45\columnwidth}
        \centering
        \includegraphics[width=\linewidth]{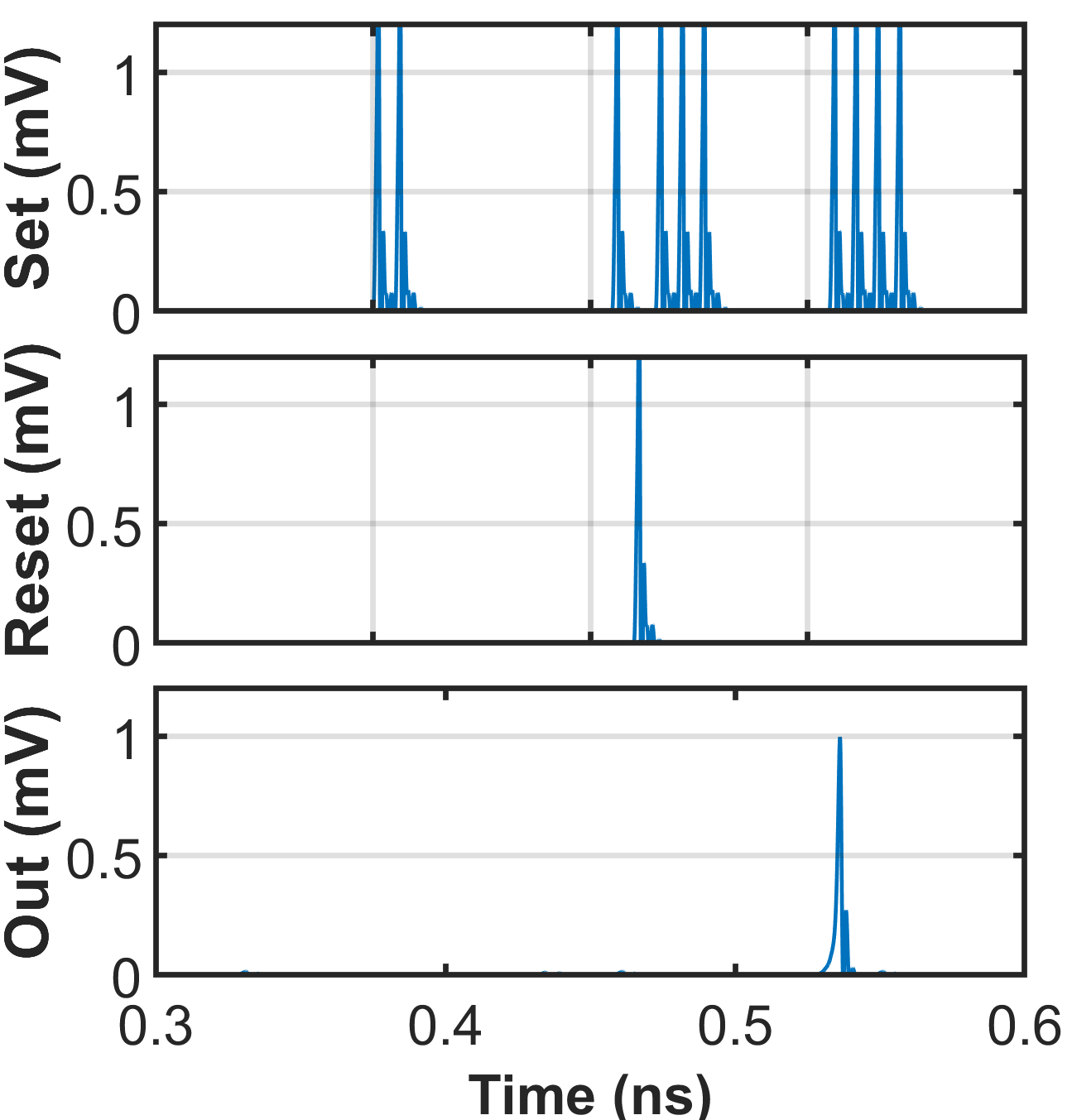}
        \captionsetup{font=small}
        \caption{Simulations result of a TU with two RTFF, which means the threshold value of four.}
        \label{fig:sim_th4}
        \vspace{-2mm}
    \end{minipage}
    \hfill
\end{figure}

TU was implemented for threshold values of 2 and 4. Simulation results for these values are shown in Fig. \ref{fig:sim_th2} and Fig. \ref{fig:sim_th4}, respectively.
 
Observing the simulation results, it is evident that a neuron has a threshold of 2 when it fires only if the input equals two or more. Similarly, a neuron has a threshold of 4 when it fires only if the input equals or exceeds four. The reset signal puts the circuits in the ground state.
\vspace{-2mm}

\subsection{Threshold Adjustment Unit}
\noindent
The second integral component of the circuit is referred to as the TAU. This unit is responsible for initializing the thresholding mechanism by loading a predetermined number of pulses. 
It configures the circuit up to a specific point, which is restricted by the maximum threshold value allowed by the hardware constraints of the thresholding mechanism.
Therefore, this essential component necessitates the storage of multiple SFQ pulses to facilitate the initial data-loading process into the thresholding unit. Moreover, it sequentially mandates the provision of data. To address these requirements, we employ the M-NDRO (Multifluxon Non-Destructive Read Out) unit, a suitable design choice for this purpose, as referenced in \cite{mndro}.

\begin{figure}[!htb]
\centering
  \includegraphics[width=0.8\linewidth]{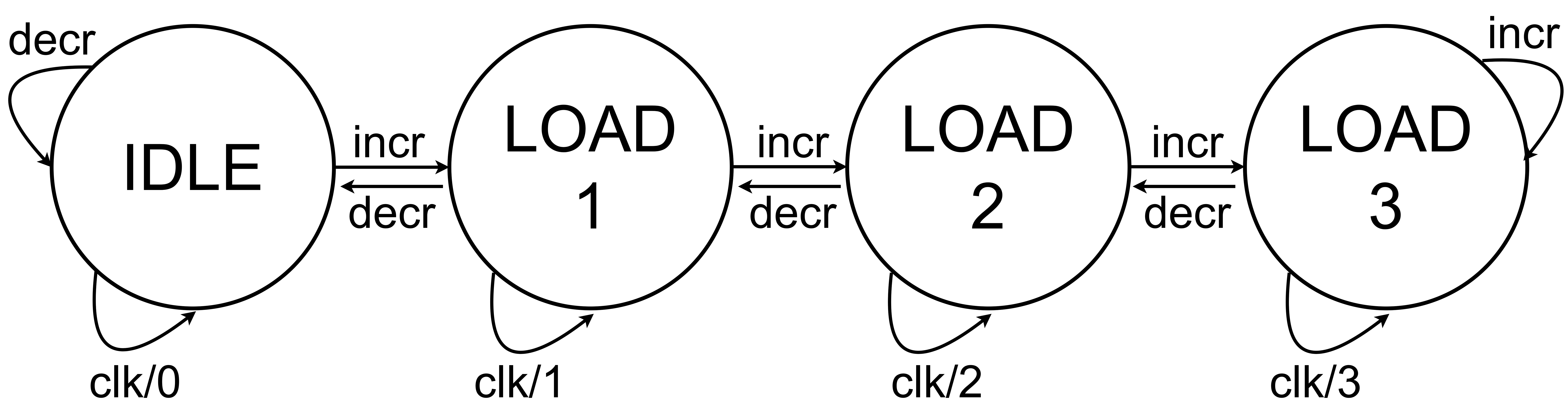}
  \caption{State Machine of TAU. When the circuit is in an Idle state, the clock signal generates no output, whereas the decrement signal maintains the idle state. Each increment signal advances the machine to a higher state; progressively more SFQ pulses are generated until the last state is reached.}
  \label{fig:sm_tau}
  \vspace{-2mm}
\end{figure}

The proposed M-NDRO unit in \cite{mndro} can store up to 3 SFQ pulses, offering dedicated increment, decrement, clock, and output ports. Notably, the data stored within the M-NDRO unit remains undisturbed until the increment or decrement signal arrives. The state machine of the TAU is given in Fig.~\ref{fig:sm_tau}. It has four states whereby the increment and decrement signals cause state transitions. When the clock signal arrives, the state machine generates 1 SFQ pulse in LOAD 1, 2 SFQ pulses in LOAD 2, and 3 SFQ pulses in LOAD 3.
\vspace{-3mm}

\begin{figure}[!h]
\centering
  \includegraphics[width=0.8\linewidth]{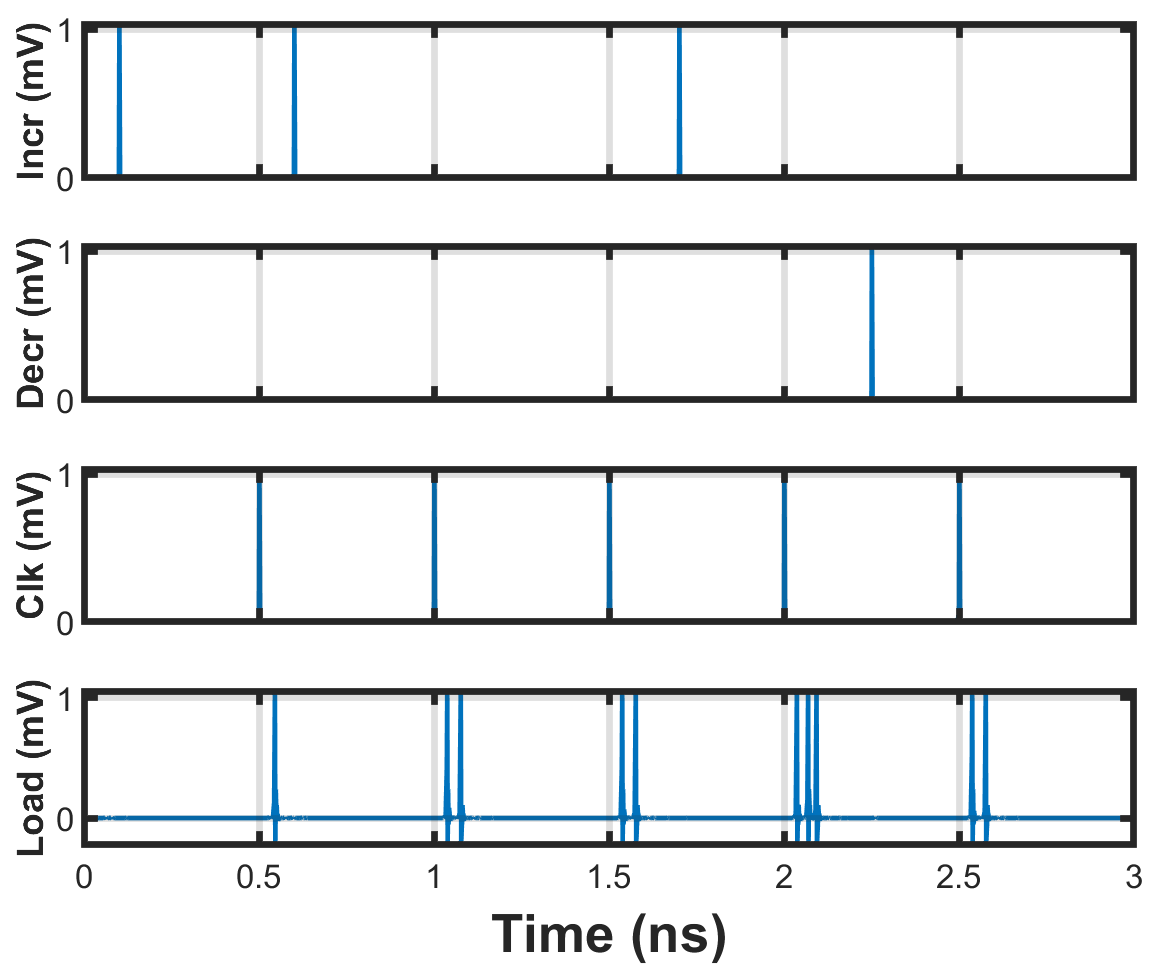}
  \caption{Simulation result of a TAU demonstrates the increment and decrement function. With the Incr signal, the number of SFQ pulses at each clock increases. The Decr signal will reduce the generated pulses at clock signal arrival.}
  \label{fig:tau_sim}
  \vspace{-2mm}
\end{figure}

The circuit has a non-destructive structure. Thus, only one load operation is sufficient to maintain the output. This attribute ensures data preservation for each received clock signal, enhancing the efficiency and reliability of the storage mechanism.
The circuit loads the initial data sequentially, which is the suitable data form for TU. The inherent non-destructive readout characteristic of the M-NDRO unit reinforces the integrity and stability of the data storage process, complementing the circuit's dynamic threshold tuning mechanism. The simulation result of TAU is reported in Fig. \ref{fig:tau_sim}.

The maximum storage capability of TAU limits the full data load. In this state-of-the-art representation, since the M-NDRO can store up to 3 SFQ pulses, the maximum value of the load data is 3. Since M-NDRO has a scalable structure for storing more SFQ pulses, this maximum value can be increased if needed.

\subsection{Arbiter Circuit}
\begin{figure}[!h]
\centering
  \includegraphics[width=0.6\linewidth]{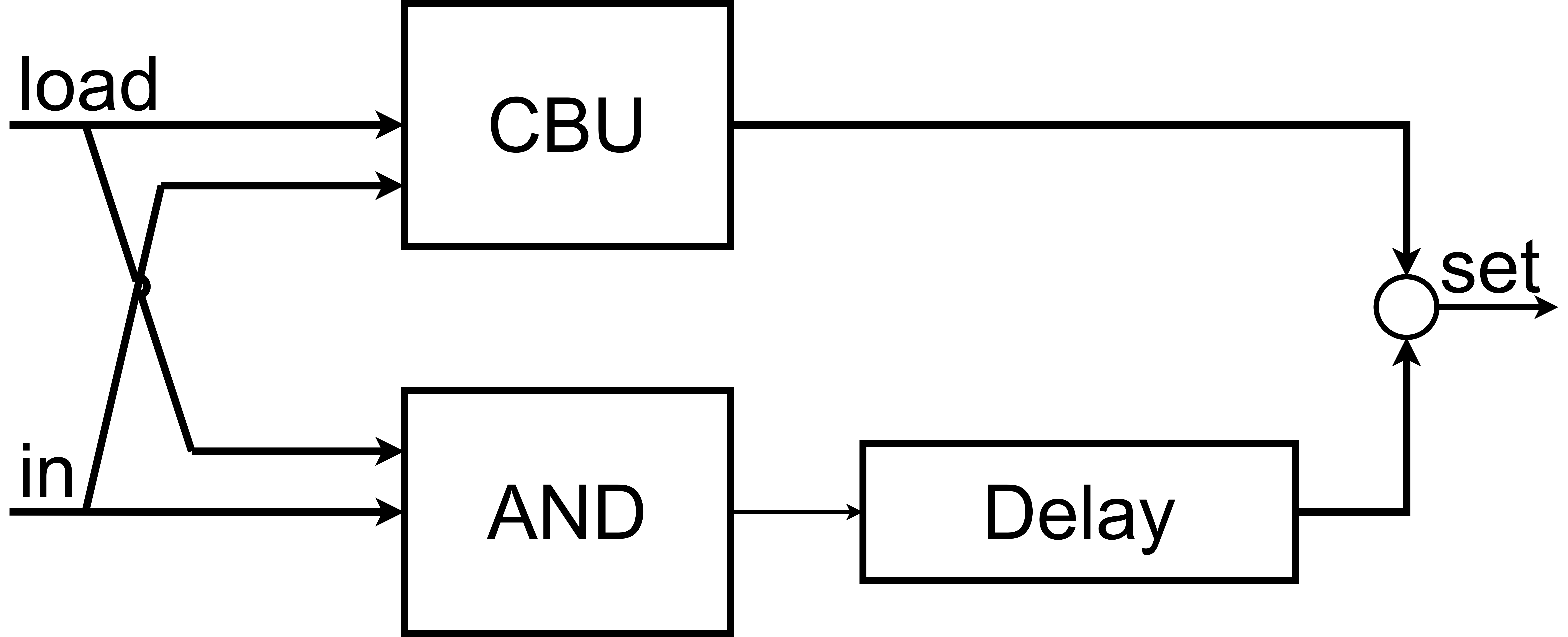}
  \caption{Arbiter block diagram shows the lossless CBU, which merges the Load and input and sends them to the set signal. When two pulses arrive in a short timing window, the CBU generates just one pulse; however, in that case, asynchronous AND will generate a pulse and apply it to the output.}
  \label{fig:arb_bd}
  \vspace{-2mm}
\end{figure}

\noindent
We have designed a novel Arbiter circuit to mitigate the risk of data loss upon precise timing specifications. The block diagram of this circuit is shown in Figure \ref{fig:arb_bd}. In this configuration, when both the load data and the input data arrive within a specific time window, the CBU generates one pulse, while concurrently, the asynchronous AND cell yields the other pulse. Subsequently, these two pulses can be combined with the delayed version of the AND cell's output, generating the set data for the TAU. The delay value depends on the maximum number of input pulses and may be adjusted for the input window. This order of the pulses and data flow safeguards against potential data loss, enhancing the reliability and robustness of the system, especially in scenarios with stringent timing constraints.

  \vspace{-5mm}
\section{Simulation and Results}
\begin{figure}[!htb]
\centering
  \includegraphics[width=0.8\linewidth]{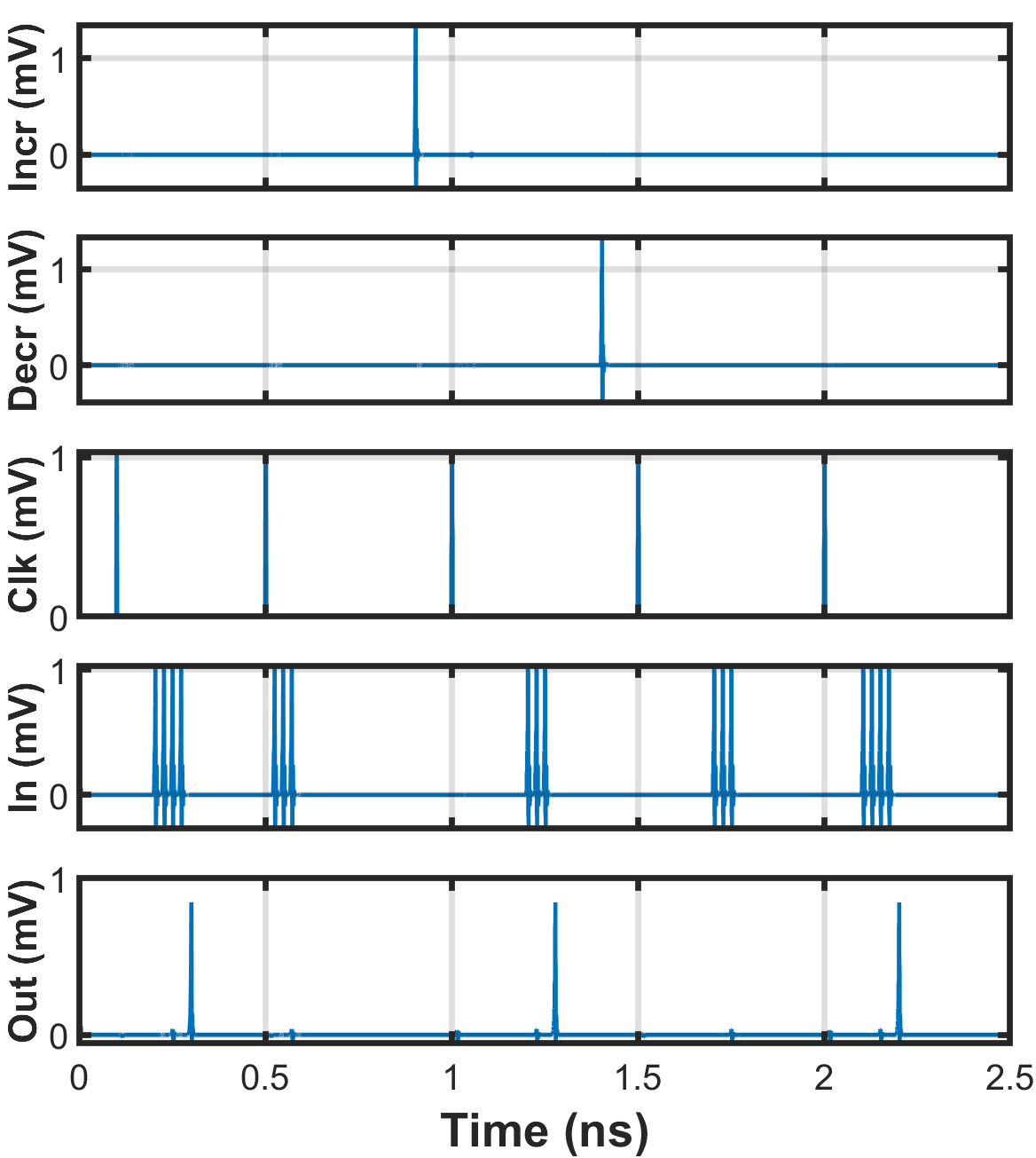}
  \caption{Simulation result for threshold values four and three. The default value is four. With an incoming Incr pulse, the threshold value becomes three and with the Decr pulse, the threshold value goes back to four.}
  \label{fig:th4th3}
  \vspace{-2mm}
\end{figure}

\noindent
To observe the circuit's functionality, circuit simulations were performed under different scenarios. In the first simulation reported in Fig. \ref{fig:th4th3}, the threshold value is changed from four to three and then switched to four again. There is no load value in the first two clock cycles, so the threshold value is set to its maximum value of four. Therefore, after four SFQ pulses in the first cycle, the neuron fires and generates an SFQ pulse. In the second clock cycle, three SFQ pulses arrive at the TU but cannot exceed the threshold value, so there is no output. Before the third clock cycle, the increment signal generates a load pulse. After this point, the threshold value is updated from four to three. Therefore, three input pulses cause an output in the neuron. In the next cycle, the decrement signal arrives and eliminates the load value. Thus, the threshold value is assigned as four again and behaves accordingly.

\begin{figure}[!htb]
\centering
  \includegraphics[width=0.8\linewidth]{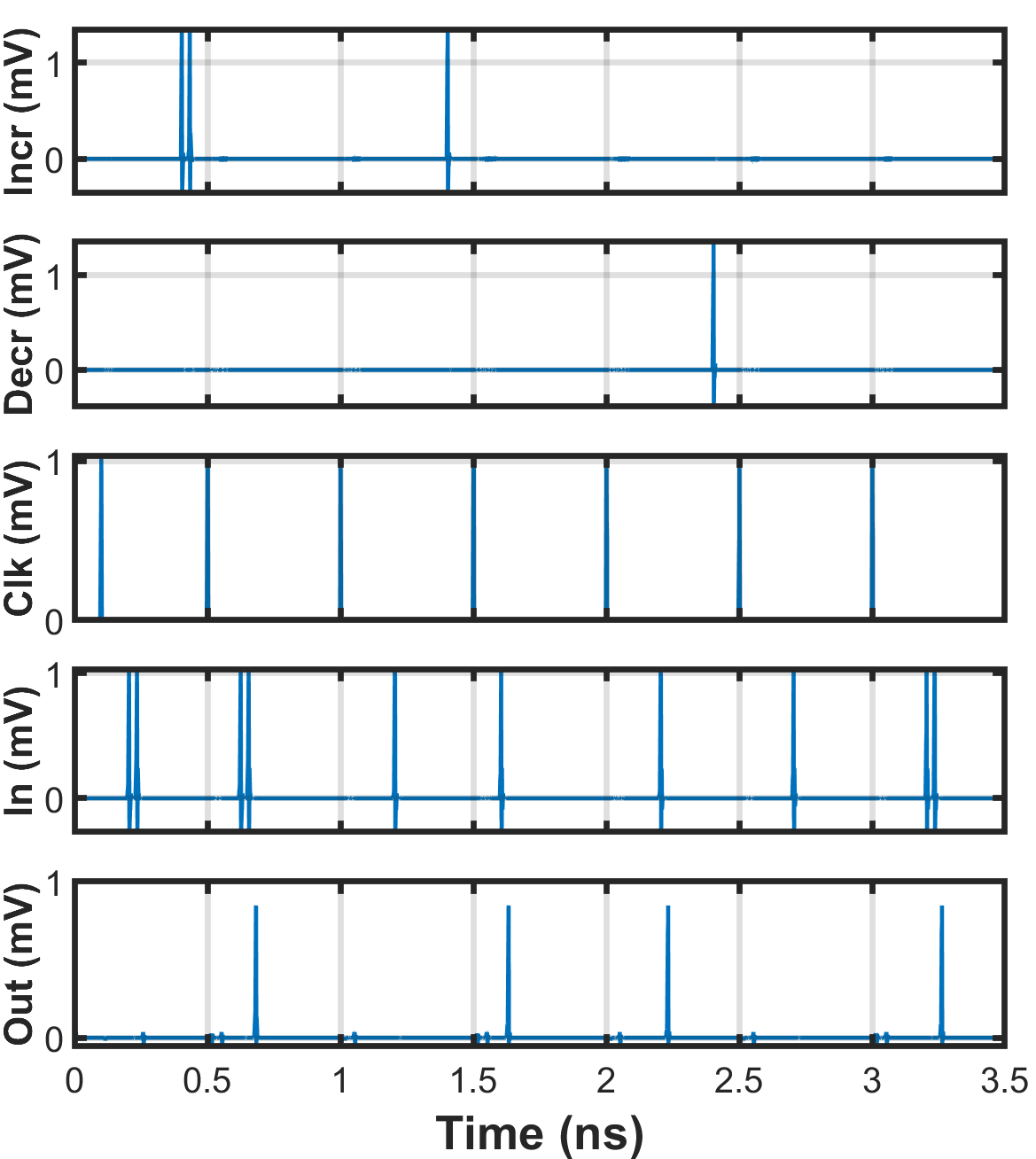}
  \caption{Simulation result for threshold values two and one.}
  \label{fig:th2th1}
  \vspace{-4mm}
\end{figure}

In the second simulation shown in Fig. \ref{fig:th2th1}, the threshold change is observed from two to one and back to two again. At the first clock cycle, the threshold value of the circuit has the maximum threshold value of four. Hence, giving two SFQ pulses from the input port doesn't trigger the neuron. Before the 2nd clock cycle, two SFQ pulses were given to the increment port to generate two load pulses. After the clock signal, it loads the data and sets the threshold value as two. Because of this, after the second clock signal, an SFQ pulse is generated. At the 4th clock, another increment signal adjusts the neuron's threshold again. It changes the TAU state to LOAD3 and inserts 3 SFQ pulses into the TU. Therefore, the threshold value is set to 1 at this point. 
One SFQ pulse is enough to meet the trigger point, and the neuron fires. When the decrement signal arrives, it changes the TAU state from LOAD3 to LOAD2 and again sets the threshold as two. After this point, one SFQ pulse is insufficient to trigger the neuron. 
As seen from the simulations, the proposed circuit supports changing the neuron's threshold value repeatedly. In this case, each threshold change takes 40ps. The simulations adjust the threshold multiple times to observe the different scenarios. However, only one initial threshold should be enough for the inference neural network implementations. 

An alternative approach is investigated given the possible hardware overhead involved in incorporating a threshold-adjusting mechanism into each circuit. Rather than making individual adjustments to the threshold values of each neuron, the idea is to modify the threshold for an entire network layer or specific kernels. This approach provides a more practical solution to mitigate the risk of excessively high hardware costs.
To evaluate the efficacy of the proposed approach, experiments were conducted using the MNIST dataset. By systematically changing the threshold values of neurons in different layers, we observed a positive impact on the accuracy of the neural network. For example, in the 128-96-96-10 network, by changing only the second layer's threshold value from four to two while keeping the first, third, and fourth layers stay at the thresholds 4, 2, and 2, respectively, the accuracy is increased from 95.9\% to 97.1\%. This outcome underscores the benefits of adopting the layer threshold to enhance network performance.

For some implementations, the threshold value is too low for the input if a neuron or number of neurons fire in every round. In these cases, the adjustable threshold mechanism gives us the advantage of increasing the thresholds to make the neuron values suitable for the network. Furthermore, some neurons never fire, which are called "dead neurons." This means the specified threshold value is high for the network. Again, with the help of the new neuron architecture, we can adjust the correct threshold and make the dead neuron functional. 
\vspace{-2mm}

\section{Conclusion}
\noindent
We presented an on-chip trainable neuron design where the threshold values of the neuron circuit can be increased or decreased at run time, depending on the specific network or its applications. The Threshold Unit, Threshold Adjustment, overall Neuron Circuit design, and comprehensive simulation results were provided. and discussed The threshold adjustment time is 40ps, and the overall circuit's margins are 20\%.

\textbf{Acknowledgments:}
This work has been funded by the National Science Foundation under the DISCoVER Expedition: with grant number 2124453.


\end{document}